\documentclass{article} 

\usepackage{iclr2027_conference,times}

\usepackage{amsmath,amsfonts,bm}

\def\eqref#1{equation~\ref{#1}}

\def\1{\bm{1}}

\DeclareMathAlphabet{\mathsfit}{\encodingdefault}{\sfdefault}{m}{sl}
\SetMathAlphabet{\mathsfit}{bold}{\encodingdefault}{\sfdefault}{bx}{n}

\usepackage{multirow}
\usepackage[hidelinks]{hyperref}
\usepackage{cleveref}

\usepackage{url}
\usepackage{graphicx}
\usepackage{wrapfig}
\usepackage{tcolorbox}
\usepackage{enumitem}
\usepackage{booktabs}
\usepackage{xcolor}
\usepackage{tabularx}

\newtcolorbox{diagnosticpromptbox}{
  colback=black!3,
  colframe=black,
  boxrule=1.3pt,
  arc=2mm,
  left=9pt,right=9pt,top=8pt,bottom=8pt,
  before skip=8pt,after skip=8pt,
  fontupper=\normalsize\itshape
}
\title{When Words Speak Louder than Images: Towards Understanding Language Bias in Vision–Language Models}

\author{
Yizhou Fang$^{\clubsuit}$,
Siyue Chen$^{\diamondsuit}$,
Zimo Qi$^{\heartsuit}$,
Zhiyu Xue$^{\spadesuit}$,
Xi Chen$^{\dagger}$,
Guangliang Liu$^{\ddagger}$\thanks{
\parbox[t]{0.92\linewidth}{
\raggedright
Email: \texttt{<y276fang@uwaterloo.ca>}.\\
Correspondence to Guangliang Liu: \texttt{<liugua@iu.edu>}.\\
Dataset:
\href{https://huggingface.co/datasets/Eric555zz/human-cross-modal-coverage-annotations}
{\texttt{Eric555zz/human-cross-modal-coverage-annotations}}
}
}
\\[5pt]
$^{\clubsuit}$University of Waterloo
\qquad
$^{\diamondsuit}$Independent Researcher
\\
$^{\heartsuit}$Johns Hopkins University
\qquad
$^{\spadesuit}$University of California, Santa Barbara
\\
$^{\dagger}$Nanyang Technological University
\qquad
$^{\ddagger}$Indiana University
}

\newcommand{\GL}[1]{\textcolor{red}{\textbf{[Comment:} #1]}}
\newcommand{\bi}[1]{\textbf{\textit{#1}}}
\iclrfinalcopy
\begin{document}

\maketitle
\lhead{Preprint.}
\begin{abstract}
Despite substantial progress across downstream applications, vision–language models (VLMs) remain susceptible to \textit{language bias}, often prioritizing linguistic cues over visual evidence and consequently producing incorrect predictions.
Prior studies have proposed various approaches to understanding and mitigating language bias in VLMs, yet their findings often conflict due to the difficulty of \textit{tracing how language bias propagates within black-box VLMs}.
Building on the word completion task, we trace how language bias propagates through VLM inference by (1) proposing a \textit{diagnostic} framework that decomposes the inference process into four distinct yet interdependent stages to trace the propagation of language bias; and (2) examining how two key factors underlying language bias, i.e.,~\textit{linguistic priors} and \textit{cross-modal coverage}, evolve across these stages and ultimately give rise to incorrect predictions.
The linguistic prior captures the strength of statistical bias induced by the language model component of a VLM and represents the origin of language bias, whereas cross-modal coverage measures the extent to which linguistic cues cover the visual content.
By decomposing inference into four stages and characterizing the interplay between linguistic priors and cross-modal coverage across these stages, we propose a systematic framework for tracing the propagation of language bias throughout the inference process; and uncover the underlying mechanism of language bias by revealing the interplay between linguistic priors and cross-modal coverage.
\end{abstract}

\hypersetup{hidelinks}

\section{Introduction\label{sec:intro}}

\begin{wrapfigure}{r}{0.5\linewidth}
    \centering
    \includegraphics[width=0.99\linewidth]{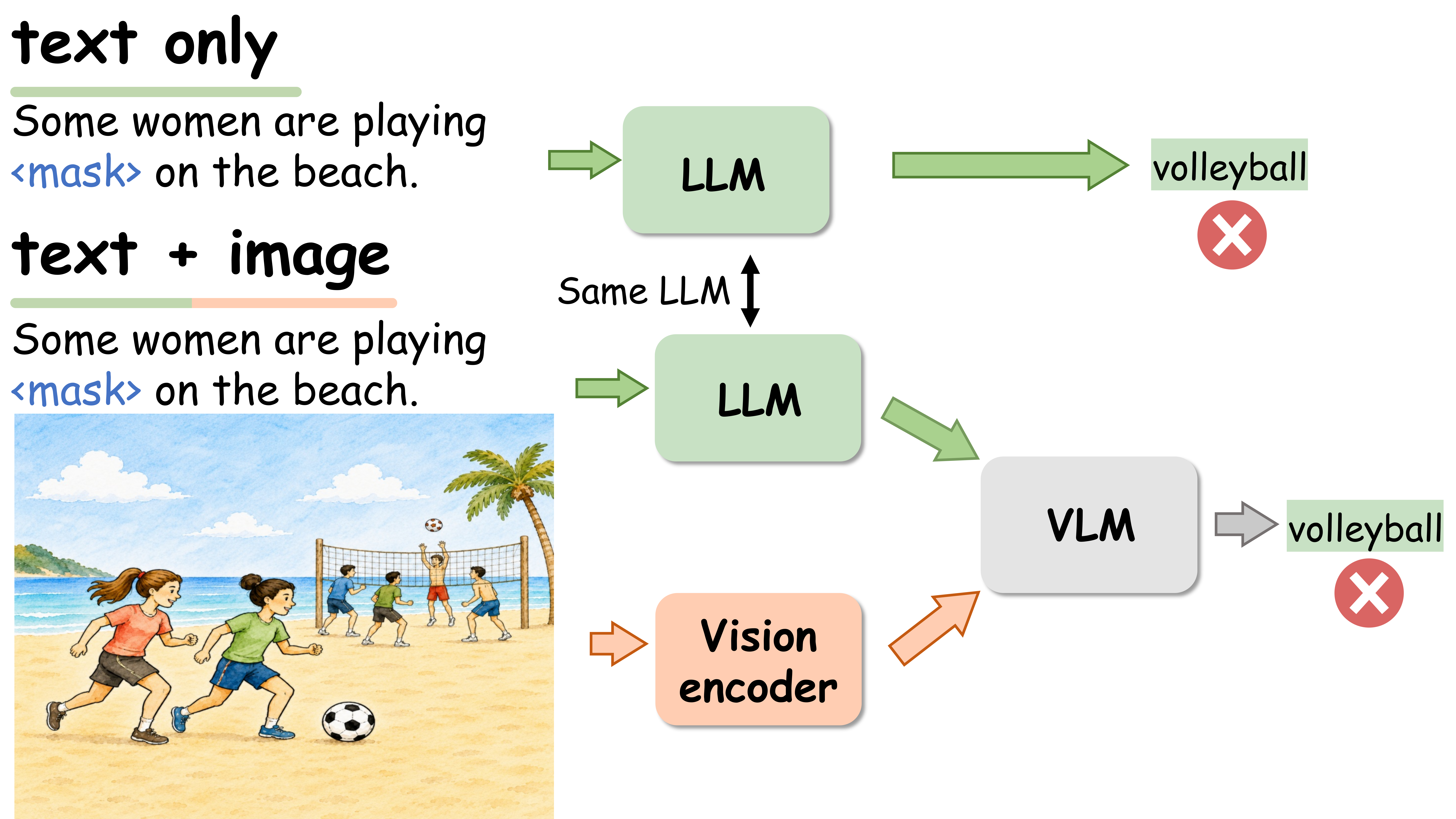}
    \caption{\small An example of language bias in VLMs: predictions from the underlying LLM can dominate the output, even when they conflict with the visual input.}
    \vspace{-10pt}
    \label{fig:intro}
\end{wrapfigure}

Vision–Language Models (VLMs) integrate a powerful Large Language Model (LLM) with a vision encoder to support tasks such as image captioning, visual question answering, and visually grounded dialogue.
However, this architectural design can also introduce \textit{language bias}, causing models to prioritize linguistic cues over visual evidence and potentially produce incorrect predictions~\citep{liu2024insight,deng2025blind}.
For example, as illustrated in Figure~\ref{fig:intro}, given the caption
``\emph{Some women are playing \rule{1em}{0.4pt} on the beach}.'' a VLM may predict
\emph{volleyball} for the blank, consistent with a strong linguistic
preference induced by the surrounding text.
However, the image shows that the women are playing \emph{soccer}, while
the men are playing \emph{volleyball}.
Understanding why linguistic cues can override task-relevant visual
evidence is therefore central to reliable multimodal understanding.

There have been various studies in understanding and mitigating the language bias.
One line of research \textit{attributes biased predictions to specific architectural components}. 
For example, prior studies associate biased predictions with insufficient attention to visual information and show that amplifying attention to image tokens can mitigate hallucinations \citep{liu2024paying,yin2025clearsight}.
However, other studies find that manipulating visual attention does not necessarily eliminate VLMs' tendency to favor language bias~\citep{yu2026dismantling,ortu2025when}.
Another line of research hypothesizes that \textit{language bias stems from weak associations between visual and linguistic cues}, and seeks to mitigate such bias by strengthening these cross-modal associations, reporting improvements in hallucination mitigation for visual reasoning~\citep{chen2023shikra,wu2026postalign}.
But this approach fails in other tasks \citep{geigle2024does,wu2026postalign}. 
To understand the mechanisms underlying language bias, a prominent line of research leverages \textit{counterfactual data} by constructing visual content that contradicts linguistic cues and examining how VLMs respond to such conflicting multimodal inputs~\citep{liu2024insight,luo2025probing}.
However, this line of research relies heavily on the assumption that the multimodal data-generating function underlying VLMs can adequately account for counterfactual data, an assumption that remains unverified~\citep{lin2023counterfactual,zhang2024what,jeong2026multimodal}.
More discussion about related works is available in Appendix~\ref{app:relatedworks}.

These seemingly conflicting findings, together with the difficulty of selecting appropriate analysis tools, stem from the \textbf{\textit{black-box nature of VLMs}}, making it challenging to trace how language bias propagates throughout the inference process~\citep{feng-etal-2023-pretraining}.
Toward understanding the mechanisms underlying language bias, this paper traces its propagation by \textbf{(1)} proposing a diagnostic framework that decompose the inference process into four distinct yet interdependent stages, providing a structured framework for tracing how language bias propagates throughout inference; and \textbf{(2)} focusing on two fundamental factors without relying on assumptions about specific architectural components.
In addition, we focus on the \textit{word completion} task. As illustrated in Figure~\ref{fig:intro}, this task requires VLMs to predict a single word in response to a textual prompt, conditioned on the provided visual input.
This is because the target word is typically a noun or adjective, which carries substantial lexical-semantic information~\citep{baker2003lexical,baker2017lexical}, thereby facilitating our analysis of language bias.

\begin{figure*}[t]
    \centering
     \vspace{-15pt}
    \includegraphics[width=0.9\linewidth]{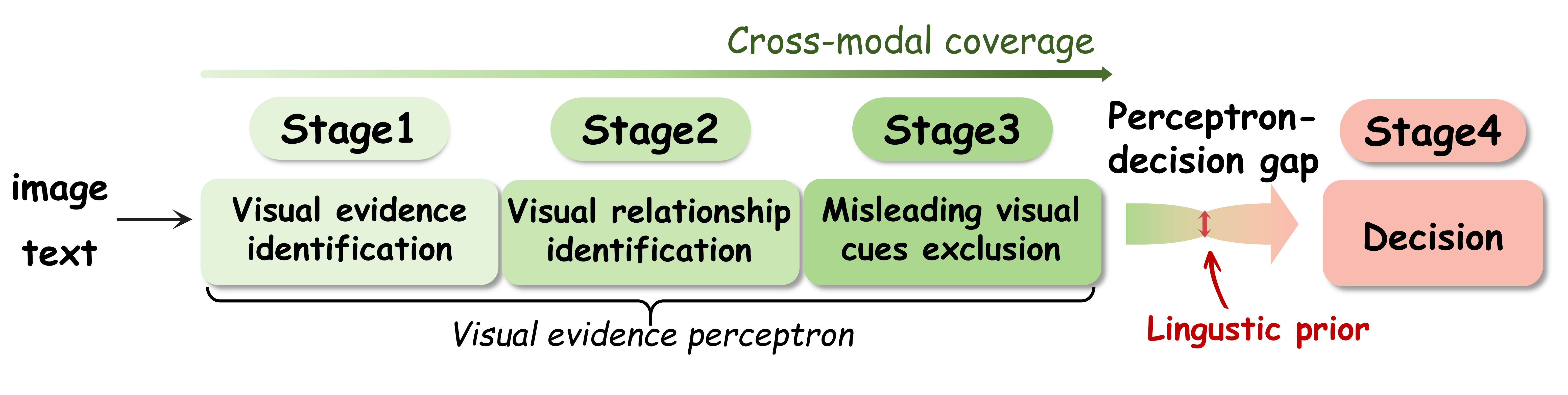}
    \vspace{-10pt}
    \caption{\small \textbf{Overview of the proposed four-stage diagnostic framework, illustrating how the interplay between cross-modal coverage and linguistic priors gives rise to language bias in VLMs.}
The framework decomposes VLM inference into visual evidence identification (Stage~1), visual relationship identification (Stage~2), misleading visual cue exclusion (Stage~3), and final decision making (Stage~4). Across the first three stages, increasing cross-modal coverage enables the model to progressively extract and integrate visual evidence. 
}
    \vspace{-20pt}
    \label{fig:pipeline}
\end{figure*}

As illustrated in Figure~\ref{fig:pipeline}, this paper traces how language bias  propagates throughout VLM inference by: 
\textbf{(1)} proposing \textit{\textbf{a diagnostic framework}} that decompose the inference process into four distinct yet interdependent stages: \text{visual evidence identification}~\citep{radford2021learning,jia2021scaling,kamath2021mdetr,li2023evaluating}, \text{visual relationship identification}~\citep{yuksekgonul2023when,liu-etal-2023-visual,pmlr-v267-hou25c,chen2025spatial}, \text{misleading visual cue exclusion}~\citep{mousi-etal-2026-correct,sun2026distractions,xie2025hope}, and \text{decision}~\citep{nooralahzadeh2026arbitration,wang2026knowing}, thereby enabling the analysis of where language bias emerges during prediction. These stages serve as a functional diagnostic decomposition of the task rather than a claim that VLMs internally execute inference through the same sequential process; 
and \textbf{(2)} identifying two key factors underlying language bias, \textit{\textbf{linguistic prior}} and \textit{\textbf{cross-modal coverage}}, and characterizing their interplay with our diagnostic framework.
The linguistic prior captures the strength of statistical bias induced by the LLM component within the VLM, representing the \textit{origin} of the language bias.
Cross-modal coverage measures the extent to which linguistic cues describe the visual content in a model-agnostic manner, thereby providing a means to evaluate the \textit{cross-modality interactions} between the two modalities.
By focusing on these two factors, we avoid the challenge of making assumptions about which architectural components within black-box VLMs contribute to language bias, while our empirical observations demonstrate the effectiveness of these two factors in understanding such bias.

Building on this setting, our empirical experiments and statistical analyses elucidate the mechanisms underlying language bias through a diagnostic framework and the interplay between two key factors. Our main findings are:

\noindent\textbf{Diagnostic Framework: Four-Stage Decomposition of the Inference Process.}
The four-stage decomposition provides a unified framework for tracing language bias and reconciling conflicting findings in prior studies. \textbf{(1) Beyond association:} successful inference requires not only visual–linguistic associations but also the identification of relevant visual relationships and the exclusion of misleading cues, explaining the limited effectiveness of association-based mitigation. \textbf{(2) Beyond attention:} the need to exclude misleading cues further shows that greater visual attention alone cannot prevent VLMs from relying on the linguistic prior.

\noindent\textbf{Interplay between Linguistic Prior and Cross-Modal Coverage.}
Their interplay across the four stages reveals how linguistic priors diminish the effects of cross-modal coverage and ultimately lead to language bias. \textbf{(1) Cross-modal coverage facilitates progression:} higher cross-modal coverage is associated with progression to later inference stages. \textbf{(2) Linguistic priors create an perceptron–decision gap:} they can diminish the benefits of cross-modal coverage before the decision stage, allowing correctly recognized visual evidence to be overridden in the final decision. 

\textbf{Organization.} \S~\ref{sec:preliminary} introduces the preliminaries for uncovering the mechanisms underlying language bias, including the diagnostic framework, linguistic priors, and cross-modal coverage.
\S~\ref{sec:mechism} presents a mechanistic analysis of language bias.
\S~\ref{sec:discussion} discusses the key findings and explores potential approaches to mitigating language bias.
Finally, \S~\ref{sec:conclusion} concludes the paper.

\newcounter{example}


%
\section{Preliminary}
\label{sec:preliminary}
In this section, we establish the preliminaries for analyzing the mechanisms underlying language bias. Specifically, we first introduce the dataset and task formulation, then present a diagnostic framework that decomposes the inference process into four stages (\S~\ref{sec:four_stage_framework}), and finally formalize the notions of linguistic prior and cross-modal coverage (\S~\ref{sec:linguistic_priors_metrics} and \S~\ref{sec:cross_modal_coverage_metrics}).



We use the dataset developed by~\citet{ma2023world}, which, as illustrated in Figure~\ref{fig:intro}, requires VLMs to perform a word-completion task based on multimodal inputs.

\textbf{Notations.} 
Let $x_v$ denote the image, $x_l$ the text containing a blank, and $w^{\mathrm{gold}}$ the ground-truth completion.
Given the multimodal input $(x_v,x_l)$, a VLM $f$ parameterized by $\theta$ produces a completion: $
\hat{w}^{\mathrm{vl}} = f_\theta(x_v,x_l).
$
Given the text alone, $f$ produces a text-only prediction:
$
\hat{w}^{\mathrm{l}} = f_\theta(x_l).
$ 
We use $P_\theta^{\mathrm{l}}(w\mid x_l)$ to denote the probability
assigned to a candidate completion $w$ under the text-only setting.
We further use $p_{\mathrm{ling}}$ to denote the linguistic-prior score
and $p_{\mathrm{cov}}$ to denote cross-modal coverage.

\subsection{A Four-Stage Diagnostic Framework}
\label{sec:four_stage_framework}
\begin{figure*}[t]
    \centering
    \includegraphics[width=1.0\linewidth]{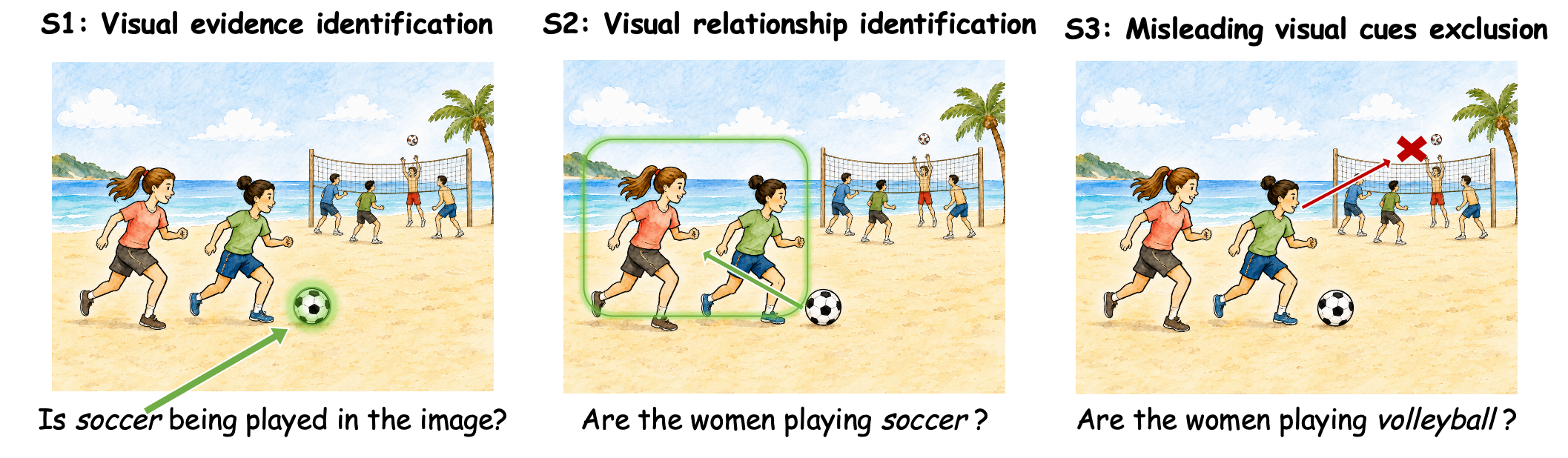}
    \caption{\small \textbf{Visualization of what each stage (the first three stages) measures in our diagnostic framework.}     \textit{S1: Visual evidence identification.} Assesses whether VLMs can identify the visual evidence corresponding to the ground-truth completion.
\textit{S2: Visual relationship identification.} Assesses whether VLMs can correctly associate the ground-truth visual evidence with its relevant visual cues.
\textit{S3: Misleading visual cue exclusion.} Assesses whether VLMs can reject misleading visual cues. \textbf{Notably}, in our diagnostic framework, \textit{S4: Decision} corresponds directly to the VLM's prediction process and is therefore not included in the figure.}
    \label{fig:visual_stages}
\end{figure*}

Examining how language bias propagates during VLM inference is challenging due to the black-box nature of VLMs.
Motivated by prior work on the mechanistic analysis of VLMs, we introduce a diagnostic framework that decomposes the inference process into four stages. 
We provide empirical evidence showing that our framework can effectively distinguish multimodal examples exhibiting language bias from those without such bias.
Figure~\ref{fig:visual_stages} illustrates what each stage measures in our four-stage diagnostic framework, as detailed below:

\textbf{S1: Visual evidence identification.}
S1 measures to what extent VLMs can adequately identify the visual evidence of the ground-truth noun completion, e.g., soccer. Such cross-modal association is widely regarded as a fundamental prerequisite for effective visual-language grounding~\citep{ma2023world,vong2024grounded}.

\textbf{S2: Visual relationship identification.}
S2 measures the extent to which VLMs can correctly identify the visual relationships associated with the ground-truth answer, e.g., the relationships between women and soccer and between men and volleyball. Prior studies have demonstrated that accurately capturing such visual relationships is crucial for effective vision-language understanding and reasoning~\citep{pmlr-v267-hou25c,liu-etal-2023-visual}.

\textbf{S3: Misleading visual cue exclusion.}
Even when VLMs correctly identify the relevant visual relationships, their predictions can still be misled by distracting or conflicting visual cues. In particular, correctly recognizing a true visual relationship does not necessarily imply that the model can reject a plausible but false alternative~\citep{sun2026distractions,mousi-etal-2026-correct}. 
Building on S2, S3 evaluates whether VLMs can identify and resist misleading visual cues that may interfere with an otherwise correct visual relationship. For example, in Figure~\ref{fig:intro}, volleyball serves as the misleading visual cue.

\textbf{S4: Decision.} 
We further introduce a dedicated \textit{decision} stage to characterize how VLMs translate multimodal evidence into a final prediction. This distinction is motivated by mechanistic studies showing that constructing task-relevant internal representations and effectively leveraging those representations for prediction are distinct processes~\citep{geva2023dissecting,neo2025interpreting}. In other words, a model may form the correct internal representation without successfully translating it into the correct prediction.
In our implementation, the decision stage corresponds directly to the VLM's final prediction of the target word completion given the multimodal input.

As illustrated in Figure~\ref{fig:pipeline}, we regard the first three stages associated with visual cues as the \textit{\textbf{visual evidence perception}} process, which characterizes how VLMs perceive and extract evidence from visual inputs. We then focus on the decision stage, investigating how the perceived visual evidence can be leveraged to prevent the propagation of language bias.

\subsection{EVALUATION OF VLM PERFORMANCE AT EACH STAGE}
Following prior work on prompt-based probing of multimodal models
~\citep{salin2022visionlanguage,cao2022prompt,zhao2024what,
zhou2025focus,pmlr-v267-hou25c}, we use natural language probes to evaluate VLM performance at each stage of our diagnostic framework. 

\begin{figure}[ht]
\vspace{-6pt}
\centering

\begin{tcolorbox}[
    width=0.85\linewidth,
    colback=gray!8,
    colframe=black,
    boxrule=0.5pt,
    arc=1pt,
    left=4pt,
    right=4pt,
    top=3pt,
    bottom=3pt,
    boxsep=0pt
]
\footnotesize

\begin{minipage}[t]{0.485\linewidth}
\vspace{0pt}
\raggedright

\textbf{S1: Visual Evidence Identification}\par
Is \emph{soccer} being played in the image?

\vspace{2pt}
\textbf{S2: Visual Relationship Identification}\par
Q1: Are the women playing \emph{soccer}?\par
Q2: Are the men playing \emph{volleyball}?

\end{minipage}
\hfill
\begin{minipage}[t]{0.485\linewidth}
\vspace{0pt}
\raggedright

\textbf{S3: Misleading Visual Cue Exclusion}\par
Q1: Are the women playing \emph{volleyball}?\par
Q2: Are the men playing \emph{soccer}?

\vspace{2pt}
\textbf{S4: Decision}\par
The original multimodal input.

\end{minipage}

\end{tcolorbox}

\refstepcounter{example}
\label{ex:probe_example}

\vspace{1pt}
\noindent
\footnotesize
\textbf{Example~\theexample:}
\textbf{Stage-wise probing questions} for the example in
Figure~\ref{fig:intro}.

\vspace{-10pt}
\end{figure}

\paragraph{Probing questions.}



Example~\ref{ex:probe_example} illustrates the probing questions for the example shown in Figure~\ref{fig:intro}. The probing questions are designed to be straightforward and directly derived from the definition of each stage.
For S1 (visual evidence identification), we directly probe whether the ground-truth word is visually present in the image. Both S2 and S3 contain two probing questions. For S2 (visual relationship identification), the questions assess whether the VLM can identify the correct visual relationship associated with the ground-truth completion. In contrast, S3 probes potentially misleading visual relationships, for which the expected answer is \emph{No}, allowing us to examine whether visually salient but irrelevant evidence biases the model’s reasoning. Finally, for S4, we directly evaluate the VLM’s output given the complete multimodal input.
The complete prompt templates and additional examples are provided in
Appendix~\ref{app:diagnostic_qa_examples}.

\textbf{Validation of the diagnostic framework.} To further validate the four-stage design of our diagnostic framework, Table~\ref{tab:diagnostic_validation} compares stage-wise performance across four VLMs for cases with and without language bias. In both conditions, the text-only model produces an incorrect completion. In \emph{biased} cases, the VLM retains this linguistically driven incorrect prediction even after receiving the image, indicating the presence of language bias. In \emph{no-bias} cases, the VLM instead uses the visual evidence to correct the initial prediction, indicating the absence of language bias.
This provides a rigorous and stringent criterion for determining whether a case exhibits language bias.

\begin{table*}[t]

    \centering
    \setlength{\tabcolsep}{5pt}
    \renewcommand{\arraystretch}{1.08}
    \begin{tabular}{lccc|ccc}
    \hline
    \multirow{2}{*}{Model}
        & \multicolumn{3}{c|}{Biased (\%)}
        & \multicolumn{3}{c}{No-bias (\%, $\Delta$)} \\
    \cline{2-7}
        & S1 & S2 & S3
        & S1 & S2 & S3 \\
    \hline

    Gemma 3 4B
        & 92.9 & 87.5 & 16.1
        & 98.7 (+5.8) & 94.4 (+6.9) & 59.8 (+43.7) \\

    Qwen2.5-VL 7B
        & 64.1 & 74.4 & 25.6
        & 78.2 (+14.1) & 89.4 (+15.0) & 57.6 (+32.0) \\

    OneVision 1.5 4B
        & 86.4 & 88.6 & 25.0
        & 93.9 (+7.5) & 95.8 (+7.2) & 52.7 (+27.7) \\

    OneVision 1.5 8B
        & 83.8 & 83.8 & 29.7
        & 95.4 (+11.6) & 97.5 (+13.7) & 61.1 (+31.4) \\

    \hline
    \end{tabular}
    \caption{\small
    \textbf{VLM performance across Stages S1–S3 under biased and unbiased conditions.}
    Values in parentheses denote the performance gap
    $\Delta S_k$, computed as the no-bias minus biased performance
    at stage $S_k$.
    S1, S2, and S3 denote Visual Evidence Identification,
    Visual Relationship Identification, and Misleading Visual Cue Exclusion,
    respectively.
    }
    \label{tab:diagnostic_validation}
\end{table*}

Across all four VLMs, \textit{\textbf{the no-bias cases consistently achieve higher performance than the biased cases at each of the three stages, providing empirical support for the effectiveness of our diagnostic framework}}. The average performance gaps are 9.8, 10.7, and 33.7 percentage points at S1, S2, and S3, respectively. This consistent stage-wise separation demonstrates that the framework effectively distinguishes cases in which VLMs successfully use visual evidence from those in which linguistic priors lead to biased predictions. The particularly large gap at S3 further indicates that misleading visual cue exclusion is a critical stage in the emergence of language bias.

\subsection{Linguistic Prior}
\label{sec:linguistic_priors_metrics}

Linguistic prior has been widely recognized as \textbf{\textit{a major source of language
bias}} in VLMs~\citep{goyal2017making,wu2023role,lin2024revisiting,
lee2025vlind,deng2025blind}.
Here, we use linguistic prior to measure the strength of statistical bias induced by the language model component of a VLM.
We define the linguistic-prior score as:
\begin{equation}
p_{\mathrm{ling}} =
\frac{P_\theta^{\mathrm{l}}(\hat{w}^{\mathrm{l}}\mid x_l)}
{P_\theta^{\mathrm{l}}(\hat{w}^{\mathrm{l}}\mid x_l)
+ P_\theta^{\mathrm{l}}(w^{\mathrm{gold}}\mid x_l)}
\label{eq:linguistic_prior}
\end{equation}
The score $p_{\mathrm{ling}}$ quantifies the text-only model's relative preference for its prediction $\hat{w}^{\mathrm{l}}$ over the ground-truth completion $w^{\mathrm{gold}}$, with larger values indicating a stronger preference for $\hat{w}^{\mathrm{l}}$.

\subsection{Cross-modal Coverage}
\label{sec:cross_modal_coverage_metrics}

Cross-modal coverage measures how much of the visual cues relevant to the ground-truth word completion is already expressed in the textual input.
However, identifying which visual cues are relevant to the ground-truth completion is non-trivial.
An image typically contains more visual information than can be captured by a textual description~\citep{tavakoli2017paying,ilinykh2018task,kreiss2022concadia,chan2023ic3}, textual descriptions are selective, verbalizing only a subset of the visual content rather than describing everything in the image.
We therefore \bi{employ human annotators to identify visual cues relevant to the ground-truth completion} and determine which of these cues are already expressed in the textual input.

\begin{figure*}[t]
    \centering
    \includegraphics[width=1.0\linewidth]{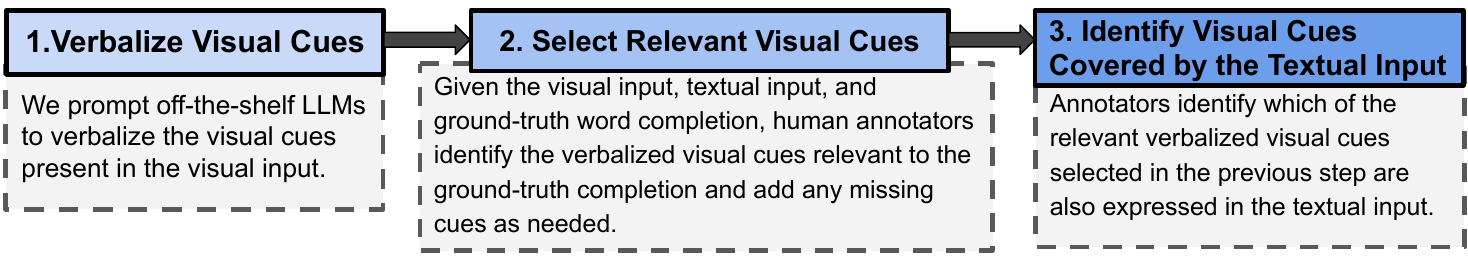}
    \caption{\small \textbf{Overview of our annotation pipeline for cross-modal coverage.} We first prompt off-the-shelf LLMs to verbalize the visual cues present in the visual input. Human annotators then identify the cues relevant to the ground-truth word completion and add any missing cues. Finally, annotators determine which of these relevant visual cues are already expressed in the textual input.}    
    \label{fig:annotation_pipeline}
\end{figure*}

To estimate cross-modal coverage, we develop a three-step annotation pipeline, as illustrated in Figure~\ref{fig:annotation_pipeline}. 
First, we prompt off-the-shelf LLMs to verbalize the visual cues present in the visual input, providing candidate cues for subsequent annotation. 
Second, given the visual input, textual input, and ground-truth word completion, human annotators identify the verbalized visual cues relevant to the ground-truth completion and add any relevant cues missed during the verbalization step. We denote the number of relevant visual cues by $N^r_V$.
Finally, annotators determine which of these relevant visual cues are already expressed in the textual input, and we denote the number of such covered cues by $N^r_L$.
This procedure allows us to distinguish between relevant visual cues that are already covered by the textual input and those that remain available only in the visual input.

\bi{Cross-modal coverage is then defined as: $p_{\mathrm{cov}} =
    N^r_L/N^r_V$.}
Thus, $p_{\mathrm{cov}}$ measures the proportion of visual cues relevant to the
ground-truth word completion that are also expressed in the textual input.
Further annotation details are provided in Appendix~\ref{app:coverage_annotation}.

\section{Mechanistic Analysis}
\label{sec:mechism}
\suppressfloats[t]
In \S~\ref{sec:preliminary}, we introduce the methodological preliminaries for analyzing the mechanisms underlying language bias. In this section, we first assess linguistic prior and cross-modal coverage as informative variables by testing their associations with final prediction correctness. Using our diagnostic framework, we then examine their roles in visual evidence perception and the final decision. Our findings are: (1) cross-modal coverage enhances visual evidence perception; and (2) linguistic priors can override the benefits of cross-modal coverage.

\paragraph{Experimental setup and backbone models.}
We evaluate five instruction-tuned VLMs from three model families: Gemma 3 (4B and 12B)~\citep{gemma3}, Qwen2.5-VL 7B~\citep{bai2025qwen25vl}, and LLaVA-OneVision 1.5 (4B and 8B)~\citep{an2025llavaonevision15}. 
All models are evaluated on the same set using human-reviewed correctness that accepts semantically valid completions consistent with both inputs.


\begin{table}[!htb]
\centering
\caption{
Associations of cross-modal coverage and linguistic prior with completion
correctness. $\beta_C$ and $\beta_P$ denote the coefficients of cross-modal coverage and linguistic prior, respectively.  For Qwen2.5-VL 7B, we additionally report results in low- and
high-prior regimes. $^{*}p<.05$, $^{**}p<.01$, and $^{***}p<.001$.
}
\label{tab:factor_regression}
\small
\begin{tabular}{llrrr}
\toprule
Model & Items & $\beta_C$ & $\beta_P$ \\
\midrule
Gemma 3 4B       & 2,839 & $1.409^{***}$ & $-1.740^{***}$ \\
Gemma 3 12B      & 2,839 & $2.209^{***}$ & $-1.910^{***}$ \\
Qwen2.5-VL 7B & 2,839 & $0.080^{\phantom{***}}$ & $-1.609^{**\phantom{*}}$ \\
OneVision 1.5 4B & 2,839 & $1.165^{***}$ & $-3.012^{***}$ \\
OneVision 1.5 8B & 2,839 & $1.082^{***}$ & $-3.713^{***}$ \\
\bottomrule
\end{tabular}
\end{table}

\subsection{Cross-Modal Coverage Enhances Visual Evidence Perception}
\label{sec:coverage_progression}
\begin{figure}[t]
    \centering
    \includegraphics[width=\linewidth]{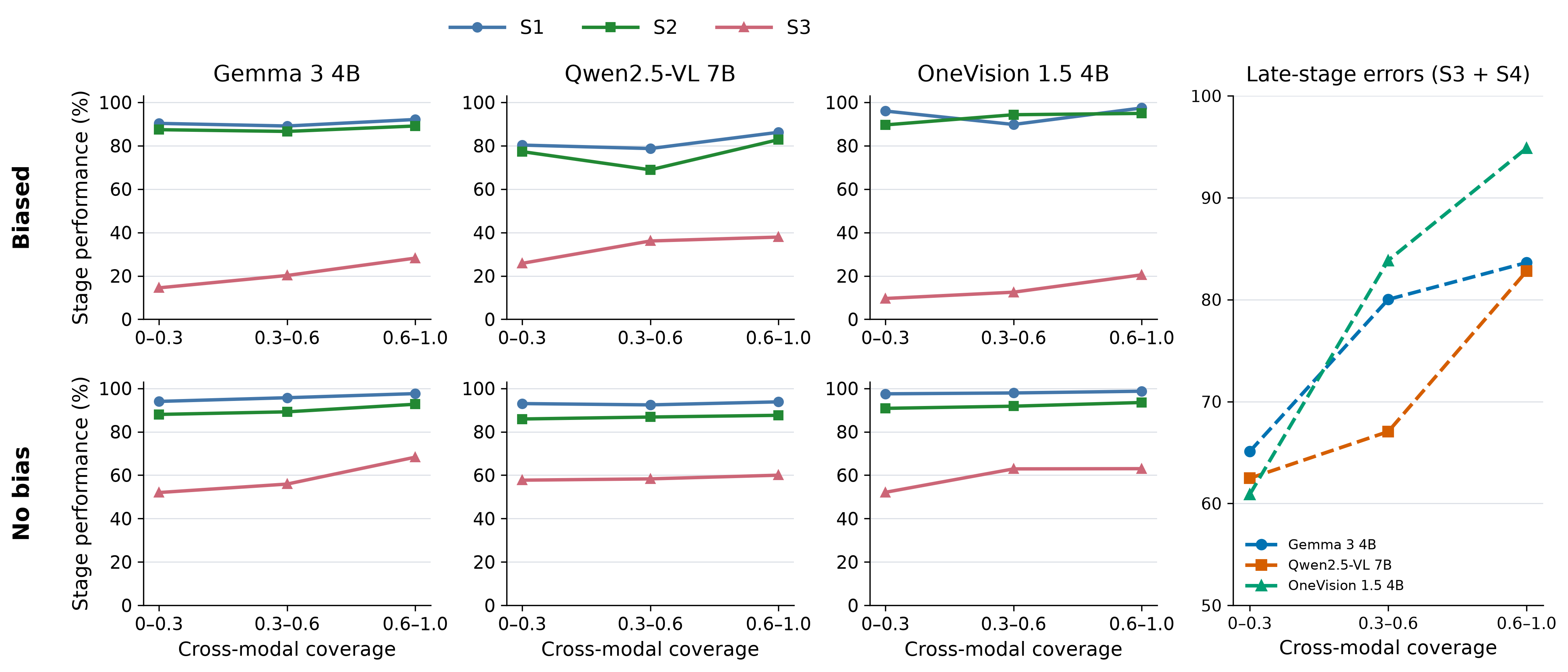}
    \caption{Diagnostic-stage performance and error distribution across cross-modal coverage intervals for one representative model from each model family. Left: S1--S3 performance on noun-target items for biased (top) and no-bias (bottom) cases across Gemma 3 4B, Qwen2.5-VL 7B, and OneVision 1.5 4B. Right: the proportion of noun-target completion errors assigned to S3 or S4 across the same coverage intervals. Results for Gemma 3 12B and OneVision 1.5 8B are provided in Appendix~\ref{app:additional_stage_results}.}
    \label{fig:late_stage_share}
\end{figure}
Prior studies have linked visually ungrounded predictions to excessive reliance on language priors~\citep{favero2024multimodal,lin2024revisiting} and insufficient use or grounding of visual evidence~\citep{favero2024multimodal,dai2023plausible}. Motivated by these two lines of work, we focus on two directly measurable variables for language bias: \emph{linguistic prior} and \emph{cross-modal coverage}.

To characterize the role of cross-modal coverage within our diagnostic framework, we \bi{examine how greater cross-modal coverage enhances visual evidence perception} and enables VLMs to progress through the first three stages. To this end, we conduct statistical analyses to examine (1) whether greater cross-modal coverage improves VLM performance across the first three stages in both biased and no-bias cases and (2) whether it facilitates VLM progression from S1 to S3.

We first examine whether cross-modal coverage and linguistic prior are systematically associated with final prediction correctness. A logistic regression jointly considering the two factors shows opposing associations with correctness: greater cross-modal coverage is positively associated with correctness for four of the five models, whereas stronger linguistic prior is negatively associated with correctness across all five models (Table~\ref{tab:factor_regression}). Together, these results provide statistical support for cross-modal coverage and linguistic prior as two informative factors associated with final prediction behavior.

We next examine how this relationship is reflected within our diagnostic framework. For this stage-wise analysis, we only include items whose ground-truth completion is a noun (\texttt{NOUN}). Noun targets allow concrete evaluation of the relevant visual concept, its relationship to the queried entity or action, and the validity of a candidate completion. Figure~\ref{fig:late_stage_share} shows that as cross-modal coverage increases, performance across the diagnostic stages generally improves, with the clearest change appearing at S3. This pattern is observed in both biased and no-bias cases, indicating that greater cross-modal coverage is associated with stronger visual evidence perception. For clarity, Figure~\ref{fig:late_stage_share} presents one representative model from each model family; results for Gemma 3 12B and OneVision 1.5 8B are provided in Appendix~\ref{app:additional_stage_results} and show the same qualitative pattern.

At the same time, among incorrect noun-target completions, the share of errors assigned to S3 or S4 increases with cross-modal coverage. Even at higher coverage levels, difficulties with misleading visual cue exclusion and the final decision remain prominent among the errors that persist. S4 cases are particularly informative: on some items, the model answers all S1–S3 probes correctly yet completes the original caption incorrectly. We examine these cases in the next section.

\subsection{The Linguistic prior overwrites visual evidence perceptron}
\label{sec:gap_analysis}

In the previous section, we showed that higher cross-modal coverage is associated with better performance on S1--S3. We now examine the role of linguistic priors in the perceptron--decision gap, which describes cases in which a VLM answers the S1--S3 probes correctly but still makes an incorrect final prediction. Specifically, we ask whether \bi{strong linguistic priors can still bias the final prediction when the model has answered the visual-evidence probes correctly}.

Before examining the counterfactual pairs, we first test whether strong linguistic priors remain associated with final-decision failures after S1--S3 have been successfully completed. As shown in Table~\ref{tab:gap_prior_split}, final accuracy is consistently lower in the very-high-prior regime across all five models, even when all S1--S3 judgments are correct, motivating a closer examination of the role of linguistic prior in the final decision.

\begin{table}[t]
\centering
\caption{Final completion accuracy among cases with correct S1--S3 judgments for $p_{\mathrm{ling}}<0.99$ and $p_{\mathrm{ling}}\geq0.99$. $\Delta$ denotes the change in accuracy from the former to the latter, in percentage points.}
\label{tab:gap_prior_split}
\small
\begin{tabular}{lccc}
\toprule
Model & $p_{\mathrm{ling}}<0.99$ & $p_{\mathrm{ling}}\geq0.99$ & $\Delta$ \\
\midrule
Gemma 3 4B       & 98.3\% ($n=58$)  & 95.6\% ($n=135$) & $-2.7$ \\
Gemma 3 12B      & 94.1\% ($n=185$) & 92.3\% ($n=750$) & $-1.8$ \\
Qwen2.5-VL 7B    & 95.7\% ($n=69$)  & 91.8\% ($n=146$) & $-3.9$ \\
OneVision 1.5 4B & 97.7\% ($n=43$)  & 89.9\% ($n=99$)  & $-7.8$ \\
OneVision 1.5 8B & 92.6\% ($n=54$)  & 90.5\% ($n=116$) & $-2.1$ \\
\bottomrule
\end{tabular}
\end{table}

To examine the effects of linguistic priors, we (1) construct adversarial textual inputs that preserve the semantics of the original inputs but induce different linguistic priors; and (2) apply these inputs to the VLMs to analyze how changes in linguistic priors affect their final decisions. Constructing such examples is challenging because the linguistic prior must be altered without changing cross-modal coverage. To address this challenge, \bi{we restrict modifications to adjectives, prepositional phrasing and verbs}, prompt off-the-shelf LLMs to generate candidate examples, and retain only those that induce a weaker linguistic prior while preserving cross-modal coverage and successful S1--S3 judgments. For models with insufficient perceptron--decision gap cases, multiple captions associated with the same image are used, with each caption reformulated and evaluated independently. One such pair is shown below:

\begin{tcolorbox}[
    colback=gray!5,
    colframe=black,
    boxrule=0.4pt,
    arc=1pt,
    boxsep=0pt,
    left=3pt,
    right=3pt,
    top=5pt,
    bottom=5pt,
    before skip=2pt,
    after skip=2pt
]
\footnotesize
\setlength{\fboxsep}{1pt}

\textbf{Original:}
``A girl is smelling a mushroom that a woman is holding
\colorbox{yellow!35}{\strut up to} her.''\\[-1pt]
\textbf{Adversarial:}
``A girl is smelling a mushroom that a woman is holding
\colorbox{red!15}{\strut in front of} her.''
\end{tcolorbox}

Both textual inputs express the same visual cues and therefore have the same cross-modal coverage, but they can induce different linguistic priors. We construct 100 reformulation pairs for each model, resulting in 500 pairs across the five evaluated models. Across these controlled examples, weakening the linguistic prior can change the final completion even though cross-modal coverage and S1--S3 judgments remain unchanged. Additional prior-counterfactual examples, including the corresponding changes in linguistic prior and final completion, are provided in Appendix~\ref{app:gap_counterfactual_examples}.

Together, the prior-stratified observation and the controlled reformulations provide complementary evidence that linguistic prior is involved in the perceptron--decision gap. Even after successful visual evidence perception, very strong linguistic priors are associated with lower final completion accuracy, and weakening the linguistic prior can change the final decision while cross-modal coverage and S1--S3 judgments remain unchanged. Our construction is intended to demonstrate that such cases can occur rather than to estimate their frequency in the full evaluation set. Because reformulation also changes the surface form of the textual input, these examples do not establish linguistic prior as the sole cause of the perceptron--decision gap.

\section{Discussion}
\label{sec:discussion}
\textbf{Language bias in current VLMs may require more challenging
settings.}
Our results suggest that language bias is increasingly difficult to observe in stronger VLMs under relatively straightforward visual--linguistic conflicts. For example, larger models can often use available visual evidence to override misleading linguistic preferences, whereas smaller models exhibit more persistent language-biased behavior. This observation suggests that future evaluations should move beyond simple modality conflicts and consider more challenging cases, where visual evidence is incomplete, ambiguous, distributed, or requires compositional reasoning. Such settings may better reveal when strong linguistic priors continue to interfere with visual evidence even in capable VLMs.

\textbf{Grounding based on statistical associations may itself introduce
bias.}
A fundamental challenge is that current vision--language grounding is largely built upon statistical associations learned from large-scale image--text datasets. These associations provide powerful priors that enable VLMs to recognize common concepts and generate fluent responses, but they can also encourage models to rely on correlations that are not aligned with the specific visual evidence in a given instance. Therefore, language bias should not be viewed only as an undesirable error introduced after training; it may partially originate from the same statistical mechanisms that enable effective multimodal learning.

\textbf{The dual role of linguistic priors in VLMs.}
Our findings imply a fundamental ``trade-off'' in mitigating language bias. The linguistic prior that contributes to biased predictions under visual conflicts is also a major source of the knowledge and generalization ability provided by large language models.
Therefore, reducing language bias by suppressing linguistic influence may also limit the ability of VLMs to leverage the broad knowledge encoded in their language models. This trade-off suggests that language bias cannot be viewed as an isolated failure mode independent of the capabilities that make VLMs effective.

\textbf{Revisiting vision--language grounding for eliminating language bias.}
Our findings suggest that reducing language bias may require going beyond statistical associations as the basis of vision--language grounding. Current VLMs largely acquire semantic alignment from large-scale image--text co-occurrences, which provides strong generalization but can also entangle useful linguistic knowledge with instance-level biases. A more fundamental direction may be to construct semantic spaces that better reflect linguistic structures and cognitive principles, allowing visual and linguistic information to be organized according to their functional roles rather than their statistical correlations alone. Such grounding mechanisms could potentially reduce language bias while preserving, or even improving, the knowledge utilization ability of VLMs.

\section{Conclusion}
This work introduces a four-stage diagnostic framework and two key factors, linguistic priors and cross-modal coverage, to study language bias in VLMs through a word-completion task. 
Progression from the first to the final stage of our diagnostic framework reflects a transition toward unbiased prediction, with the first three stages focusing on the visual evidence perceptron. 
We further provide empirical evidence demonstrating the effectiveness of the proposed diagnostic framework.
Linguistic priors capture the statistical biases introduced by the LLM component of VLMs, while cross-modal coverage measures the extent to which visual cues are represented in the textual input. 
Our findings associate greater cross-modal coverage with improved visual evidence perception and reveal a \textit{perceptron–decision gap}, whereby linguistic priors diminish the benefits of increased cross-modal coverage, ultimately leading to language bias.
\label{sec:conclusion}


\subsection*{AI use statement}

We used generative AI tools to generate candidate textual reformulations for the prior-counterfactual experiments, as described in Section~\ref{sec:gap_analysis}. We also used generative AI tools for language editing, and limited assistance with code drafting and debugging. All AI-generated experimental inputs were filtered according to the predefined experimental criteria and manually reviewed before use. AI-assisted code and analyses were checked by the authors, and all reported results and scientific claims were independently verified. Generative AI was not used to make final human annotation or correctness decisions. We take responsibility for the final content of this work, including all text, claims, and artifacts produced with the aid of generative AI.

\subsection*{Ethics statement}

This work studies language bias in vision-language models using an existing benchmark and does not involve sensitive personal data or deployment on human subjects. Our analysis is intended to improve understanding of how linguistic priors affect multimodal predictions rather than to make claims about human behavior or social groups. Human annotations are used only for evaluating task-relevant visual and textual information and model correctness. We do not release personally identifiable information or other sensitive content.

\section*{Reproducibility Statement}

We provide detailed definitions of linguistic prior, cross-modal coverage, and the diagnostic framework in Section~\ref{sec:preliminary}. The annotation procedure, additional stage-wise results, and prior-counterfactual examples are provided in the appendix~\ref{app:appendix}. We also report the evaluated models, experimental settings, and statistical analyses used throughout the experiments.

\bibliography{iclr2027_conference}
\bibliographystyle{iclr2027_conference}

\clearpage
\appendix
\section{Appendix}
\label{app:appendix}

\subsection{Related works}
\label{app:relatedworks}

\paragraph{Language Bias in Vision--Language Models.}

Behavioral studies investigate how linguistic information influences VLM predictions when it conflicts with visual evidence. Explicit textual interference can lead models to follow misleading descriptions despite contradictory images, with this preference affected by text relevance, token order, and model scale~\citep{deng2025blind}. Conflicts can also arise without explicit misleading text. When images contradict commonsense knowledge, models sometimes answer according to their parametric knowledge rather than the depicted situation~\citep{liu2024insight}.
Related work has also examined linguistic shortcuts in image--text matching. Caption likelihood can inflate retrieval performance even without visual evidence~\citep{pmlr-v235-lin24c}, while compositional evaluations reveal failures in attribute association, relational understanding, and word-order sensitivity that standard retrieval evaluations can overlook~\citep{yuksekgonul2023when}. Dedicated benchmarks further examine these behaviors through capability controls and systematic changes to the inputs.
VLind-Bench evaluates language dependence only after checking commonsense knowledge, visual perception, and commonsense bias, thereby reducing the influence of these confounding factors~\citep{lee2025vlind}.
Counterfactual and out-of-distribution evaluations contrast cases that can be answered from linguistic priors with cases that require visual discrimination, revealing substantial weaknesses when familiar linguistic associations no longer hold~\citep{liu2024insight,luo2025probing}.
Controlled image--question groups further expose language hallucination, visual illusion, and inconsistencies across related responses~\citep{guan2024hallusionbench}.
Together, these evaluations characterize not only whether predictions are incorrect, but also whether they follow linguistic cues, respond to changes in visual evidence, and remain consistent across related questions.

Mitigation methods address language bias and associated hallucinations through training, decoding, additional visual evidence, and output revision.
PostAlign combines visual grounding with textual rationales and rejection of nonexistent objects to improve fine-grained visual understanding and reduce hallucinations~\citep{wu2026postalign}.
During generation, visual contrastive decoding contrasts predictions under original and distorted images to reduce reliance on statistical biases and unimodal priors~\citep{leng2024mitigating}.
Visual evidence can also be made explicit through specialist-model outputs~\citep{li2025visualevidence} or image descriptions that guide subsequent decoding~\citep{ghosh2025vdgd}.
At the output level, LURE revises generated descriptions using signals related to object co-occurrence, uncertainty, and position in the generated text~\citep{zhou2024lure}.
However, mitigation gains depend on the evaluation setting: adding object-grounding objectives has little to no effect on hallucination in open-ended caption generation in the experiments of~\citet{geigle2024does}. We complement these evaluations by diagnosing language-biased behavior at the level of functional judgments, and then examine how these diagnostic patterns vary with linguistic prior and cross-modal coverage.

\paragraph{Mechanistic Analysis of Language and Vision--Language Models.}

Mechanistic studies examine how internal representations and computational components contribute to model predictions.

Information-flow analyses of factual recall distinguish the enrichment of subject representations from the extraction of queried attributes~\citep{geva2023dissecting}.
Circuit analyses identify interacting attention heads and connections that support specific behaviors through both manual investigation and automated discovery~\citep{wang2023interpretability,conmy2023automated}.
The reliability of such explanations is itself an active area of study:
RAVEL evaluates the disentanglement of attributes in distributed representations~\citep{huang2024ravel}, while activation-patching studies show that localization can depend on the evaluation metric and corruption method~\citep{zhang2024patching}.

Other work directly intervenes on internal computation.
Inference-Time Intervention steers activations toward directions associated with truthful answers~\citep{li2023inference}, while DoLa exploits differences between layer-wise predictions to improve factual generation~\citep{chuang2024dola}.
Related analyses also examine discrepancies between internal truth-related representations and generated responses~\citep{liu-etal-2023-cognitive}.

In multimodal models, causal tracing has been used to identify components involved in visual information transfer~\citep{basu2024understanding}, while targeted attention-head interventions can shift predictions toward visual evidence or parametric knowledge~\citep{ortu2025when}.
Corresponding methods strengthen visual attention~\citep{liu2024paying,yin2025clearsight}, modify attention patterns associated with hallucination~\citep{huang2024opera}, steer multimodal representations~\citep{liu2025vti}, or separate visual and textual routes within attention heads~\citep{cheng2026crg}.
Recent studies further show that visual information may remain encoded even when the final prediction contradicts it, and that its influence on the answer depends on where and how that information is used~\citep{nooralahzadeh2026arbitration,wang2026knowing}.

\subsection{Diagnostic Prompt Templates and Examples}
\label{app:diagnostic_qa_examples}

The diagnostic questions used in our experiments are instantiated from
shared prompt templates.
For each item, annotators identify the ground-truth concept, the competing
concept, and the visual entities and relationships associated with them.
The surface wording is adapted to the semantic type of each item, while the
functional judgment evaluated at each stage remains fixed.

\begin{table}[ht]
\centering
\small
\setlength{\tabcolsep}{4pt}
\renewcommand{\arraystretch}{1.18}

\begin{tabular}{@{}p{0.31\linewidth}p{0.63\linewidth}@{}}
\toprule
\textbf{Stage} & \textbf{Shared Prompt Template} \\
\midrule

\textbf{S1: Visual Evidence Identification} &
Is [GROUND-TRUTH CONCEPT] visually present in the image?

For activity concepts:
``Is [GROUND-TRUTH ACTIVITY] being performed in the image?''

For object concepts:
``Can you see [GROUND-TRUTH OBJECT] in the image?''

Answer with only Yes or No. \\[5pt]

\midrule

\textbf{S2: Visual Relationship Identification} &
Q1: [TRUE RELATION BETWEEN TARGET ENTITY AND GROUND-TRUTH CONCEPT]?

Q2: [TRUE RELATION BETWEEN THE OTHER ENTITY AND COMPETING CONCEPT]?

Answer each question with only Yes or No. \\[5pt]

\midrule

\textbf{S3: Misleading Visual Cue Exclusion} &
Q1: [MISLEADING RELATION BETWEEN TARGET ENTITY AND COMPETING CONCEPT]?

Q2: [MISLEADING RELATION BETWEEN THE OTHER ENTITY AND GROUND-TRUTH CONCEPT]?

Answer each question with only Yes or No. \\[5pt]

\midrule

\textbf{S4: Decision} &
Original multimodal input.
No additional diagnostic question is provided. \\

\bottomrule
\end{tabular}

\caption{
Shared prompt templates for evaluating the four stages of the diagnostic
framework.
S2 verifies the correct visual relationships, whereas S3 constructs
misleading relationships by exchanging the concepts associated with the
corresponding entities.
Bracketed fields are instantiated from each word-completion item.
}
\label{tab:diagnostic_templates}
\end{table}

Here, [GROUND-TRUTH CONCEPT] denotes the visual concept corresponding to the
correct completion.
[TARGET ENTITY] denotes the entity queried by the original completion task,
and [COMPETING CONCEPT] denotes the visually present concept that forms the
competing interpretation.
S2 instantiates propositions corresponding to the correct visual
relationships observed in the image.
S3 uses the same entities and concepts but pairs them incorrectly to form
misleading visual relationships.
For S1, the surface form is adapted to the semantic type of the ground-truth
concept, such as an object or an activity.
S4 uses the original image and incomplete caption directly, without any
additional diagnostic prompt.

\subsection{Cross-Modal Coverage Annotation}
\label{app:coverage_annotation}

\begin{figure*}[!htbp]
    \centering
    \includegraphics[width=1.0\textwidth]
    {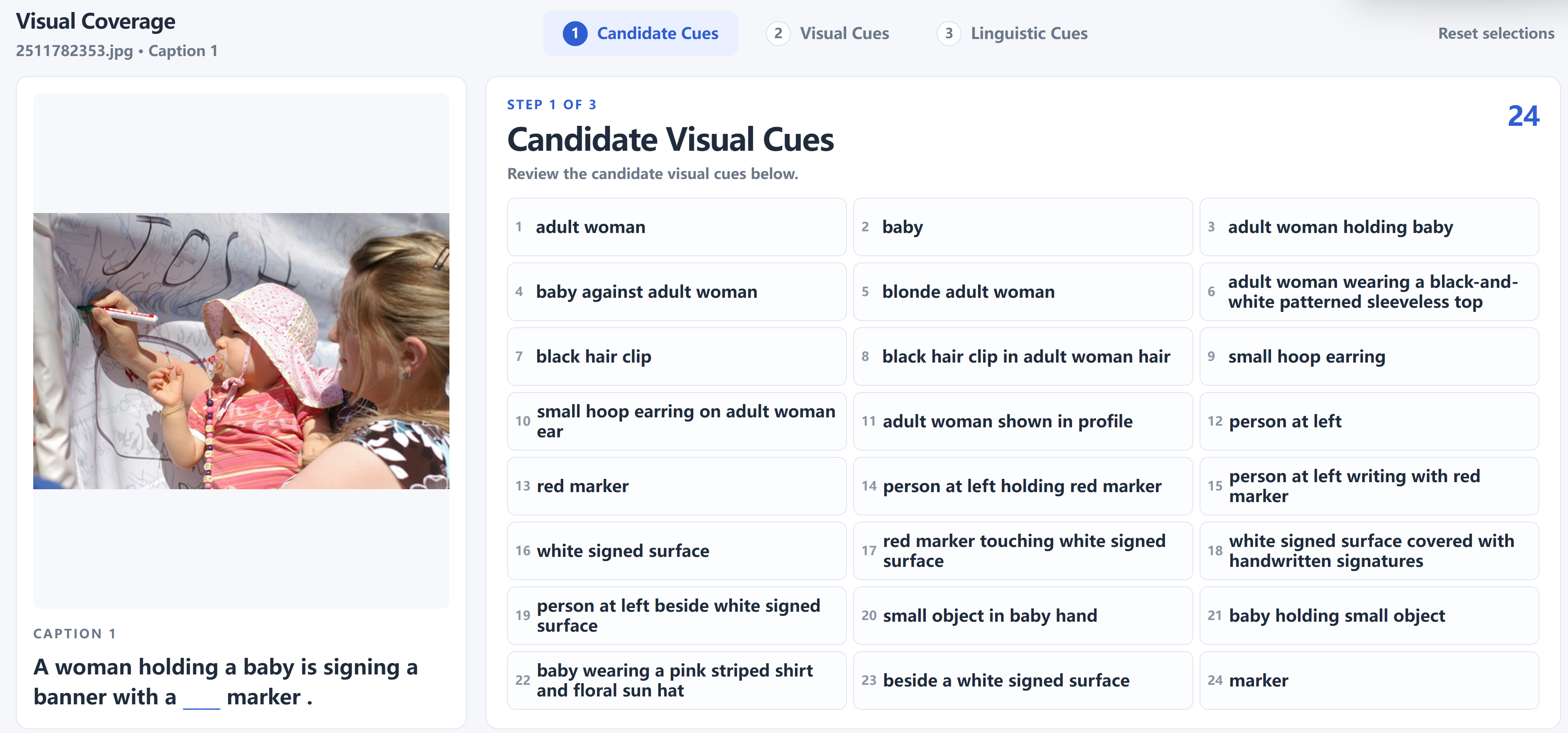}

    \caption{
    Construction and review of candidate visual cues.
    Observable entities, attributes, and relations in the example
    image are organized into 24 candidate visual cues.
    }
    \label{fig:annotation_cues}
\end{figure*}

\begin{figure*}[!htbp]
    \centering
    \includegraphics[width=1.0\textwidth]
    {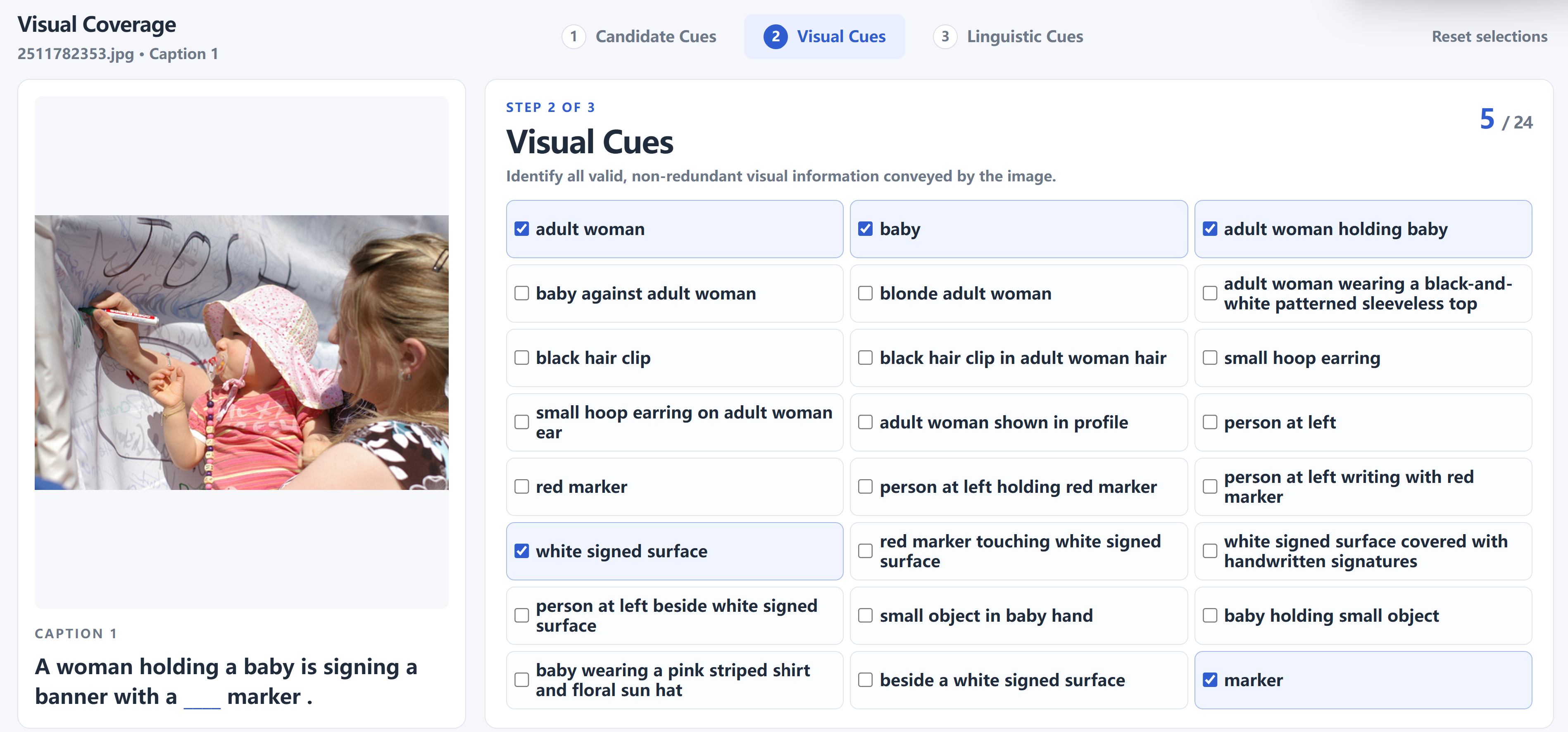}

    \caption{
    Selection of visual facts for the denominator.
    Five facts are selected from the candidate cues to form the
    visual-fact set $\mathcal{F}$:
    adult woman, baby, adult woman holding baby,
    white signed surface, and marker.
    }
    \label{fig:annotation_denominator}
\end{figure*}

\begin{figure*}[!htbp]
    \centering
    \includegraphics[width=1.0\textwidth]
    {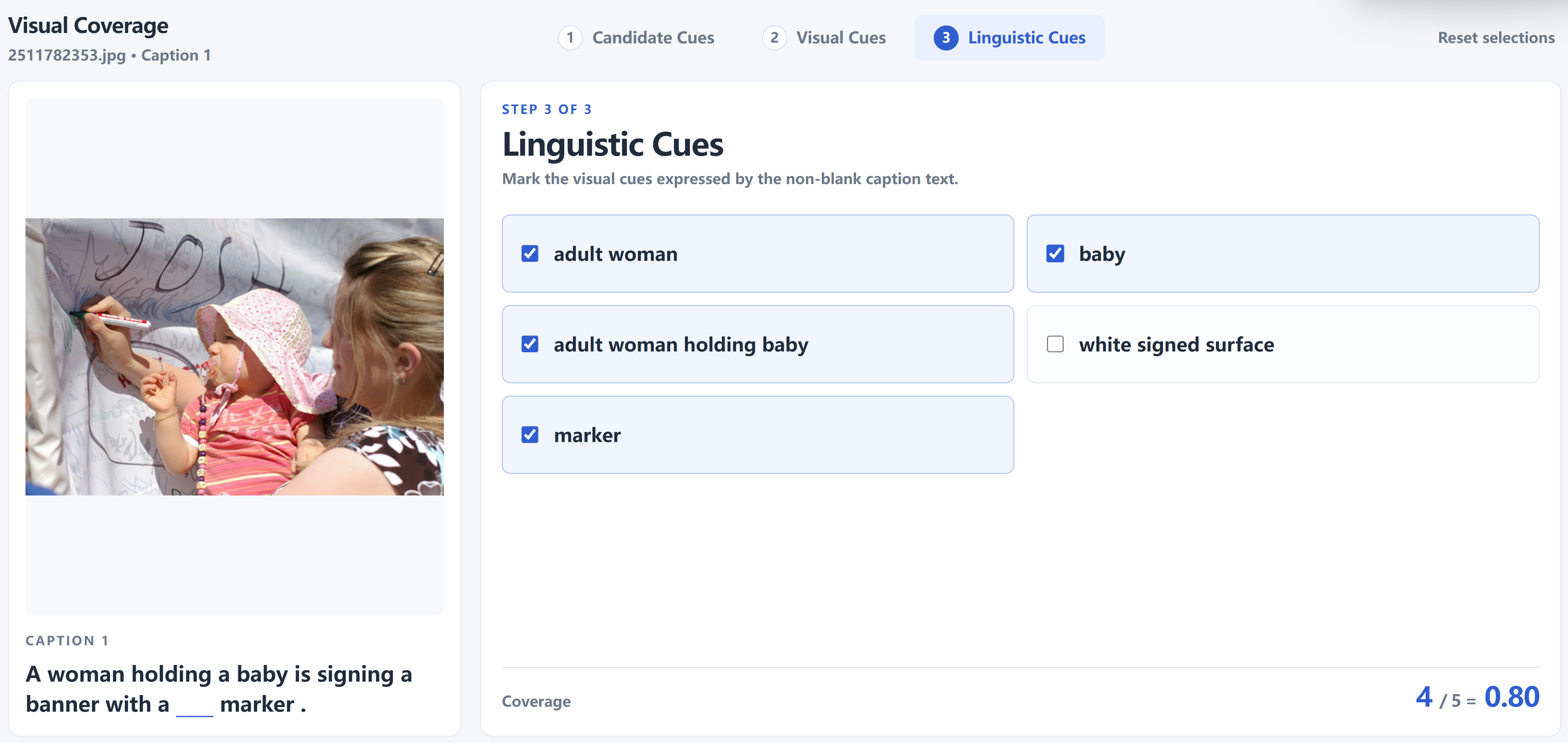}

    \caption{
    Identification of text-expressed facts for the numerator.
    Four of the five selected facts are expressed in the non-blank
    caption text, forming $\mathcal{K}(T)$ and yielding a
    cross-modal coverage of $4/5=0.8$.
    }
    \label{fig:annotation_numerator}
\end{figure*}

Figures~\ref{fig:annotation_cues}--\ref{fig:annotation_numerator}
illustrate the three-step annotation process using the same example.

\clearpage

\subsection{Additional Stage-wise Results}
\label{app:additional_stage_results}

\begin{figure}[t]
    \centering
    \includegraphics[width=\linewidth]{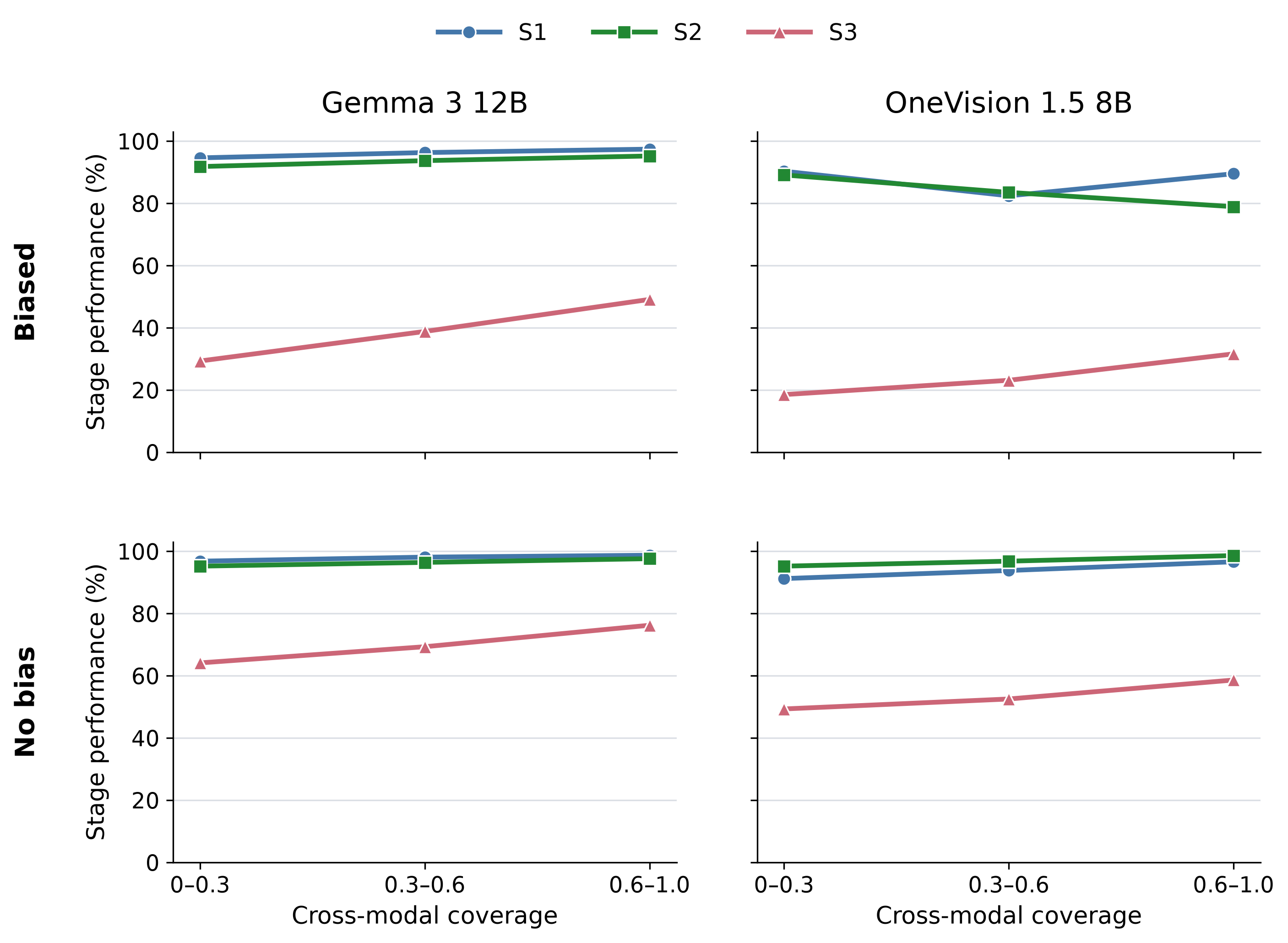}
    \caption{
    Additional stage-wise performance results for Gemma 3 12B and
    OneVision 1.5 8B across cross-modal coverage intervals.
    S1--S3 performance is reported separately for biased and no-bias cases.
    The same qualitative pattern observed in the main-text models is also
    present for these additional model sizes.
    }
    \label{fig:additional_stage_results}
\end{figure}

Figure~\ref{fig:additional_stage_results} reports the same stage-wise
performance analysis for Gemma 3 12B and OneVision 1.5 8B.
These additional model sizes show the same qualitative pattern as the
representative models reported in the main text: S1 and S2 remain relatively
stable across cross-modal coverage levels, while S3 shows a clearer improvement
as cross-modal coverage increases.
This consistency suggests that the stage-wise observations in
Figure~\ref{fig:late_stage_share} are not specific to the model sizes selected
for visualization in the main text.

\subsection{Prior-Counterfactual Examples}
\label{app:gap_counterfactual_examples}

We present ten classic counterfactual cases on distinct images.
For every case, the image, visual proposition, target, slot, gold answer,
and coverage are fixed; only a meaning-preserving reformulation lowers the
prior of the competing word.
The complete five-model witness set and validation records are provided in
the supplementary material.


\noindent
\begin{minipage}[c]{0.23\textwidth}
  \centering
  \includegraphics[
    width=\linewidth,
    height=0.14\textheight,
    keepaspectratio
  ]{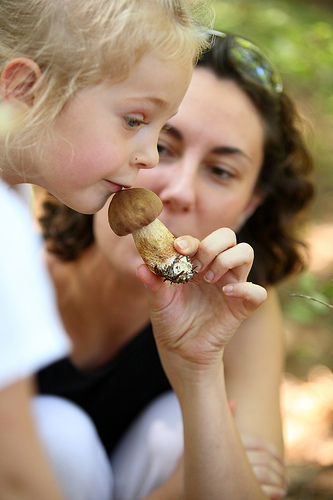}
\end{minipage}\hfill
\begin{minipage}[c]{0.74\textwidth}
  \small
  \textbf{Case 1: Qwen2.5-7B-Instruct}\\[2pt]
  \textbf{Original caption:}
  A girl is smelling a mushroom that a woman is holding up to her
  \rule{0.9em}{0.4pt}.\\[2pt]
  \textbf{Reformulated caption:}
  A girl is smelling a mushroom that a woman is holding in front of her
  \rule{0.9em}{0.4pt}.\\[2pt]
  \textbf{Prior of \emph{nose}:}
  $0.9981 \rightarrow 0.1481$
  \par\vspace{0.5em}
  \hfill
  \textbf{Answer:} \emph{nose} $\rightarrow$ \emph{face}
\end{minipage}
\par\vspace{0.45em}
\hrule
\vspace{0.65em}


\noindent
\begin{minipage}[c]{0.23\textwidth}
  \centering
  \includegraphics[
    width=\linewidth,
    height=0.14\textheight,
    keepaspectratio
  ]{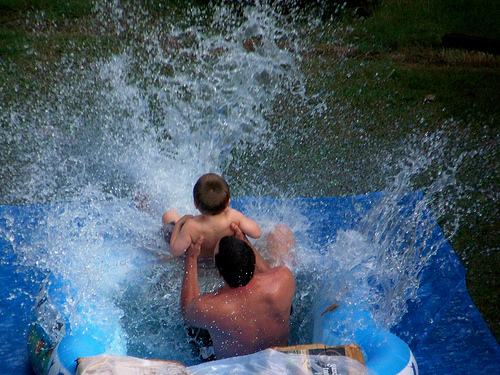}
\end{minipage}\hfill
\begin{minipage}[c]{0.74\textwidth}
  \small
  \textbf{Case 2: Qwen2.5-7B-Instruct}\\[2pt]
  \textbf{Original caption:}
  A man holds up a \rule{0.9em}{0.4pt} while sitting in a pool of water
  situated on a tarp and grassy field.\\[2pt]
  \textbf{Reformulated caption:}
  While sitting in a pool of water on a tarp and grassy field,
  a man holds up a \rule{0.9em}{0.4pt}.\\[2pt]
  \textbf{Prior of \emph{tube}:}
  $0.9526 \rightarrow 0.8671$
  \par\vspace{0.5em}
  \hfill
  \textbf{Answer:} \emph{tube} $\rightarrow$ \emph{child}
\end{minipage}
\par\vspace{0.45em}
\hrule
\vspace{0.65em}


\noindent
\begin{minipage}[c]{0.23\textwidth}
  \centering
  \includegraphics[
    width=\linewidth,
    height=0.14\textheight,
    keepaspectratio
  ]{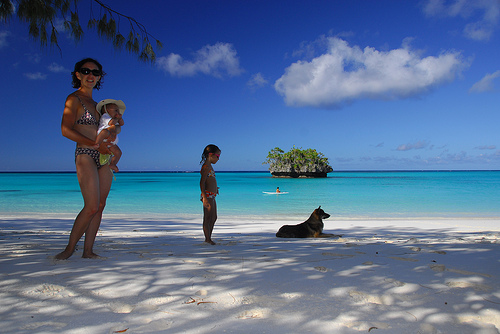}
\end{minipage}\hfill
\begin{minipage}[c]{0.74\textwidth}
  \small
  \textbf{Case 3: Qwen2.5-7B-Instruct}\\[2pt]
  \textbf{Original caption:}
  A woman in a black bikini holds a baby at the beach,
  while another little girl watches a \rule{0.9em}{0.4pt}.\\[2pt]
  \textbf{Reformulated caption:}
  A little girl watches a \rule{0.9em}{0.4pt} at the beach,
  where a woman in a black bikini holds a baby.\\[2pt]
  \textbf{Prior of \emph{paddle}:}
  $0.9399 \rightarrow 0.0067$
  \par\vspace{0.5em}
  \hfill
  \textbf{Answer:} \emph{paddle} $\rightarrow$ \emph{dog}
\end{minipage}
\par\vspace{0.45em}
\hrule
\vspace{0.65em}


\noindent
\begin{minipage}[c]{0.23\textwidth}
  \centering
  \includegraphics[
    width=\linewidth,
    height=0.14\textheight,
    keepaspectratio
  ]{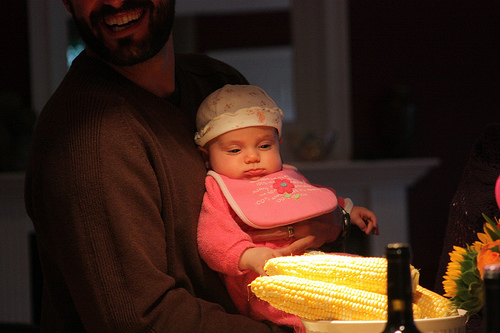}
\end{minipage}\hfill
\begin{minipage}[c]{0.74\textwidth}
  \small
  \textbf{Case 4: Qwen2.5-7B-Instruct}\\[2pt]
  \textbf{Original caption:}
  A man holding a \rule{0.9em}{0.4pt} at the dinner table.\\[2pt]
  \textbf{Reformulated caption:}
  At the dinner table, a man is holding a \rule{0.9em}{0.4pt}.\\[2pt]
  \textbf{Prior of \emph{corn}:}
  $0.9963 \rightarrow 0.8846$
  \par\vspace{0.5em}
  \hfill
  \textbf{Answer:} \emph{corn} $\rightarrow$ \emph{baby}
\end{minipage}
\par\vspace{0.45em}
\hrule
\vspace{0.65em}


\noindent
\begin{minipage}[c]{0.23\textwidth}
  \centering
  \includegraphics[
    width=\linewidth,
    height=0.14\textheight,
    keepaspectratio
  ]{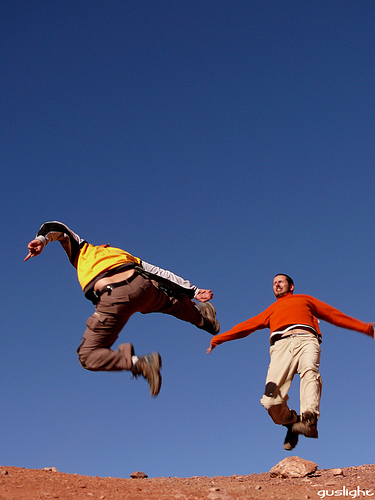}
\end{minipage}\hfill
\begin{minipage}[c]{0.74\textwidth}
  \small
  \textbf{Case 5: Qwen2.5-7B-Instruct}\\[2pt]
  \textbf{Original caption:}
  A man in a \rule{0.9em}{0.4pt} shirt and a man in an orange shirt
  jump in the air.\\[2pt]
  \textbf{Reformulated caption:}
  Men jump in the air, one wearing a \rule{0.9em}{0.4pt} shirt
  and the other orange.\\[2pt]
  \textbf{Prior of \emph{blue}:}
  $>0.9999 \rightarrow 0.8808$
  \par\vspace{0.5em}
  \hfill
  \textbf{Answer:} \emph{blue} $\rightarrow$ \emph{yellow}
\end{minipage}
\par\vspace{0.45em}
\hrule
\vspace{0.65em}


\noindent
\begin{minipage}[c]{0.23\textwidth}
  \centering
  \includegraphics[
    width=\linewidth,
    height=0.14\textheight,
    keepaspectratio
  ]{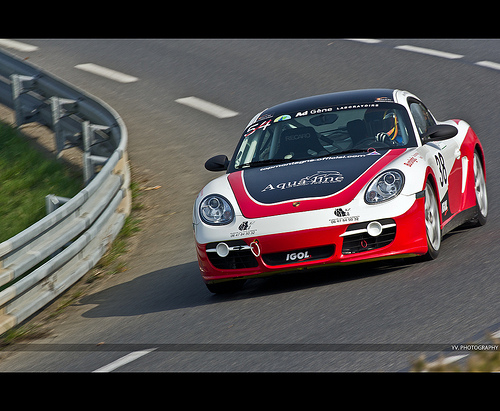}
\end{minipage}\hfill
\begin{minipage}[c]{0.74\textwidth}
  \small
  \textbf{Case 6: Qwen2.5-7B-Instruct}\\[2pt]
  \textbf{Original caption:}
  A lone red, \rule{0.9em}{0.4pt}, and black race car is being driven
  by a single driver on a racetrack.\\[2pt]
  \textbf{Reformulated caption:}
  A lone race car, being driven by a single driver, is red,
  \rule{0.9em}{0.4pt}, and black on a racetrack.\\[2pt]
  \textbf{Prior of \emph{Porsche}:}
  $>0.9999 \rightarrow 1.7\!\times\!10^{-7}$
  \par\vspace{0.5em}
  \hfill
  \textbf{Answer:} \emph{Porsche} $\rightarrow$ \emph{white}
\end{minipage}
\par\vspace{0.45em}
\hrule
\vspace{0.65em}


\noindent
\begin{minipage}[c]{0.23\textwidth}
  \centering
  \includegraphics[
    width=\linewidth,
    height=0.14\textheight,
    keepaspectratio
  ]{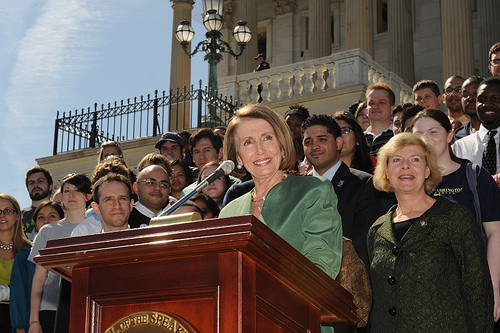}
\end{minipage}\hfill
\begin{minipage}[c]{0.74\textwidth}
  \small
  \textbf{Case 7: Qwen2.5-7B-Instruct}\\[2pt]
  \textbf{Original caption:}
  Women standing at a \rule{0.9em}{0.4pt} with a crowd and building
  in the background.\\[2pt]
  \textbf{Reformulated caption:}
  A building and a crowd appear in the background while women stand
  at a \rule{0.9em}{0.4pt}.\\[2pt]
  \textbf{Prior of \emph{steps}:}
  $0.9968 \rightarrow 0.5925$
  \par\vspace{0.5em}
  \hfill
  \textbf{Answer:} \emph{steps} $\rightarrow$ \emph{podium}
\end{minipage}
\par\vspace{0.45em}
\hrule
\vspace{0.65em}


\noindent
\begin{minipage}[c]{0.23\textwidth}
  \centering
  \includegraphics[
    width=\linewidth,
    height=0.14\textheight,
    keepaspectratio
  ]{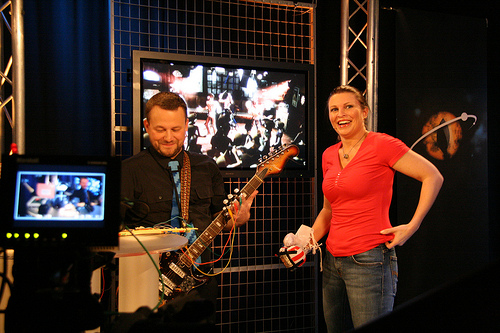}
\end{minipage}\hfill
\begin{minipage}[c]{0.74\textwidth}
  \small
  \textbf{Case 8: Qwen2.5-7B-Instruct}\\[2pt]
  \textbf{Original caption:}
  A man playing guitar and a woman wearing a
  \rule{0.9em}{0.4pt}.\\[2pt]
  \textbf{Reformulated caption:}
  A man and a woman, one playing guitar, the other wearing a
  \rule{0.9em}{0.4pt}.\\[2pt]
  \textbf{Prior of \emph{red}:}
  $>0.9999 \rightarrow 0.7163$
  \par\vspace{0.5em}
  \hfill
  \textbf{Answer:} \emph{red} $\rightarrow$ \emph{shirt}
\end{minipage}
\par\vspace{0.45em}
\hrule
\vspace{0.65em}


\noindent
\begin{minipage}[c]{0.23\textwidth}
  \centering
  \includegraphics[
    width=\linewidth,
    height=0.14\textheight,
    keepaspectratio
  ]{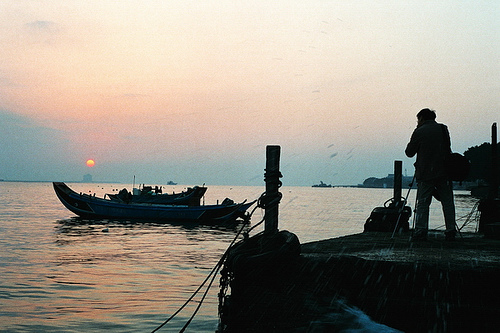}
\end{minipage}\hfill
\begin{minipage}[c]{0.74\textwidth}
  \small
  \textbf{Case 9: Qwen2.5-7B-Instruct}\\[2pt]
  \textbf{Original caption:}
  A man on a boat \rule{0.9em}{0.4pt} looking onto the water
  at a boat.\\[2pt]
  \textbf{Reformulated caption:}
  A boat is near a man on a boat \rule{0.9em}{0.4pt},
  looking at the water.\\[2pt]
  \textbf{Prior of \emph{watches}:}
  $>0.9999 \rightarrow 7.3\!\times\!10^{-7}$
  \par\vspace{0.5em}
  \hfill
  \textbf{Answer:} \emph{watches} $\rightarrow$ \emph{dock}
\end{minipage}
\par\vspace{0.45em}
\hrule
\vspace{0.65em}


\noindent
\begin{minipage}[c]{0.23\textwidth}
  \centering
  \includegraphics[
    width=\linewidth,
    height=0.14\textheight,
    keepaspectratio
  ]{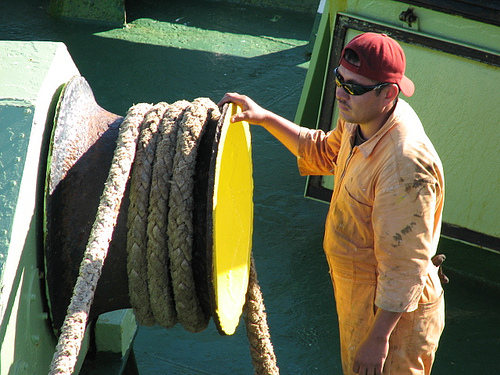}
\end{minipage}\hfill
\begin{minipage}[c]{0.74\textwidth}
  \small
  \textbf{Case 10: Qwen2.5-7B-Instruct}\\[2pt]
  \textbf{Original caption:}
  A man in an orange jumpsuit rests a hand on a very large reel of
  \rule{0.9em}{0.4pt} rope.\\[2pt]
  \textbf{Reformulated caption:}
  A very large reel holds \rule{0.9em}{0.4pt} rope,
  and a man in an orange jumpsuit rests a hand on it.\\[2pt]
  \textbf{Prior of \emph{nylon}:}
  $>0.9999 \rightarrow 6.8\!\times\!10^{-8}$
  \par\vspace{0.5em}
  \hfill
  \textbf{Answer:} \emph{nylon} $\rightarrow$ \emph{thick}
\end{minipage}
\par\vspace{0.45em}
\hrule
\vspace{0.65em}

In all ten cases, Recognition, Grounding, Relational Binding, and Exclusion
remain successful before and after reformulation.
The gallery is qualitative evidence for the controlled prior intervention
and is not a population repair-rate estimate.

\end{document}